# Large-Scale News Classification using BERT Language Model: Spark NLP Approach

*Kuncahyo Setyo Nugroho*

Department of Informatics Engineering, Faculty of Computer Science, Brawijaya University, Indonesia, ksnugroho26@gmail.com

*Anantha Yullian Sukmadewa*

Department of Informatics Engineering, Faculty of Computer Science, Brawijaya University, Indonesia, ananthayullian@gmail.com

*Novanto Yudistira*

Department of Informatics Engineering, Faculty of Computer Science, Brawijaya University, Indonesia, yudistira@ub.ac.id

The rise of big data analytics on top of NLP increasing the computational burden for text processing at scale. The problems faced in NLP are very high dimensional text, so it takes a high computation resource. The MapReduce allows parallelization of large computations and can improve the efficiency of text processing. This research aims to study the effect of big data processing on NLP tasks based on a deep learning approach. We classify a big text of news topics with fine-tuning BERT used pre-trained models. Five pre-trained models with a different number of parameters were used in this study. To measure the efficiency of this method, we compared the performance of the BERT with the pipelines from Spark NLP. The result shows that BERT without Spark NLP gives higher accuracy compared to BERT with Spark NLP. The accuracy average and training time of all models using BERT is 0.9187 and 35 minutes while using BERT with Spark NLP pipeline is 0.8444 and 9 minutes. The bigger model will take more computation resources and need a longer time to complete the tasks. However, the accuracy of BERT with Spark NLP only decreased by an average of 5.7%, while the training time was reduced significantly by 62.9% compared to BERT without Spark NLP.

## 1 INTRODUCTION

Natural language processing (NLP) is a subfield of artificial intelligence (AI) that can study human and computer interactions through natural languages, such as the meaning of words, phrases, sentences, and syntactic and semantic processing. In early, NLP research used rule-based methods to understand and reason a text. Experts manually create these rules for various NLP tasks [1]. It is complicated to manage the rules if the number of rules is large. Therefore, this approach is considered obsolete by researchers [2]. Internet development causes data to be collected easily so that a statistical learning approach is possible to resolve NLP tasks. This is known as the machine learning approach. With feature engineering, this approach brings significant improvements to many NLP tasks [3]. Meanwhile, deep learning approaches were introduced to NLP in 2012 after success in image recognition [4] and speech recognition [5]. Deep learning outperformed the other approaches with surprisingly better results.

In NLP, language modeling (LM) provides a context that differentiates similar words and phrases according to the context in which they appear. The NLP framework based on deep learning for language modeling has entered a new chapter. This is characterized by many deep learning architectures and models from which to solve NLP tasks is constantly evolving. Previously successful architecture is bidirectional LSTM (bi-LSTM) based on recurrent neural network (RNN), where the model can read the context from left to right and from right to left [6]. The main limitations of bi-LSTM are sequential, which makes the parallel training process very difficult. The transformer architecture accomplishes this by replacing the LSTM cells with an "attention" mechanism [7]. With this attention, the model can see the entire sequence of context as a whole, making it easier to practice in parallel. The Transformer has made great progress on many different NLP benchmarks. There are many transformer-based language models, including BERT[8], RoBERTa [9] GPT-2 [10] and XLNet [11].

The rise of "big data" analytics on top of NLP has led to an increasing need to ease the computational burden that processes text at scale [12]. The amount of unstructured textual data has led to increased interest in information extraction technology from academia and industry. One of the problems faced in NLP is that text has very high dimensions [13]. It takes computation capable of processing high-dimensional textual data quickly. Input data is distributed across multiple machine clusters to complete within a reasonable time. The MapReduce allows easy parallelization of large computations and uses re-execution as the primary mechanism for fault tolerance [14]. Previous research has used this concept to perform sentiment analysis tasks. The results obtained are that MapReduce can improve the efficiency of processing large amounts of text even though the performance obtained is similar to traditional sentiment analysis [15].

This research aims to study big data processing on NLP tasks based on a deep learning approach. We classify large amounts of news topics using BERT based on transformer architecture. Training BERT from scratch requires a huge dataset and takes much time to train. Therefore, we use the existing pre-trained models [8], [16].To demonstrate the efficiency of this method, we conducted extensive experiments to study our proposed approach. We use Spark NLP built on top of Apache Spark as a library that can scale the entire classification process in a distributed environment [17]. We compared the performance of the base method model with the classifier pipelines from Spark NLP. Apart from observing the model's accuracy, we also look at the computation time and computation resources used during the training and testing process.

## 2 RELATED WORK

Big data comes with an unstructured format, mainly textual data, called big text [18]. Social media has the most contribution to a big text. In addition, other online sources such as online news portals, blogs, health records, government data provide rich textual data for research. Despite the abundance of data sources, this field has attracted less attention from academia. In this section, we present literature studies carried out in the fields of deep learning for text classification and big data framework for large-scale text processing. We reviewed prior work to understand its limitations so that we can use them to refine our research.

Deep learning gives us big potential in the NLP field [19]. Many studies have contributed to text classification tasks using deep neural networks. Some successful architectures include convolutional neural network (CNN) based models, for example, VD-CNN [20] and DP-CNN [21], recurrent neural network (RNN) based models, for example, SANN [22], and attention-based models, for example, HAN [23] and DiSAN [24]. These models use pre-trained word embedding [25], [26] to improve performance in downstream tasks. Although many impressive results have been achieved, the dependent problem carries many limitations for enhancing the model's performance. Even with the development of contextualized word vectors such as CoVe [27] and ELMo [28], the model architecture still needs to be assigned in particular. Pre-training language models and fine-tuning of downstream tasks have made breakthroughs in NLP. Howard and Ruder proposed ULMFiT [29], whereas Radford et al. proposed OpenAI GPT [30] using a multi-layer transformer architecture to learn language representations of large-scale text. To solve unidirectional language representation from OpenAI GPT, Devlin et al. proposed BERT [8] using deep bidirectional representations. Compared to the previous model, BERT does not require a specific architecture for each downstream task, so this model has achieved great success in many NLP downstream tasks [31].

Hadoop is a MapReduce platform used for distributed processing. One of the Hadoop framework's major problems is that it transforms any computation as a MapReduce job [12]. In NLP, this would require re-implementation of each NLP pipeline, so it is ineffective. Apache Spark addresses this problem by extending Hadoop ecosystem with a parallel computational programming model, including resilient distributed datasets (RDDs) and learning algorithms [32]. Next, Xiangrui et al. introduced MLib[1] as a machine learning library running on Spark [33]. Research from Jian et al. analyzed the Spark framework by running a machine learning instance using MLib and highlighting Spark's advantages [34]. Spark is also used as a distributed framework for solving NLP tasks such as sentiment analysis [35], [36], and document classification [37]. Their research results show that Spark has a speed advantage in large text processing. As deep learning models have successfully in NLP, there is a need to implement pre-trained models and scale large data with distributed use cases. John Snow Labs[2] developed Spark NLP as a library built on top of Apache Spark and Apache MLib that provides an NLP pipeline and pre-trained models [17]. The library offers the ability to train, customize and save models so they can be run on clusters, other machines, or stored.

## 3 METHODOLOGY

### 3.1 Dataset

We use a corpus of news articles from the AG dataset [38]. It contains 1 million news articles that have been gathered from more than 2000 from ComeToMyHead news sources. This dataset includes 120,000 training samples and 7,600 test samples. We only use the description as a sample and category as the label. Each sample is a short text divided into four labels.

---

[1] https://spark.apache.org/mllib.
[2] https://nlp.johnsnowlabs.com.

## 3.2 BERT

BERT is a deep learning architecture that can be used for downstream NLP tasks. The architecture consists of a stacked encoder layer from the transformer [7]. There are two main steps in BERT: pre-training and fine-tuning [8]. During pre-training, BERT is trained in a large unlabeled corpus with two unsupervised tasks: masked language model (MLM) and next sentence prediction (NSP) to produce a pre-trained model. For fine-tuning, the model is initialized with the pre-trained parameters, and all the parameters are fine-tuned using labeled data for specific tasks such as classification.

We can assume the pre-trained model as a black box with H = 768 shaped vectors for each input token in a sequence. Sequences can be one sentence or a pair of sentences separated by a [SEP] token and begin with a [CLS] token. For classification task, we added an output layer to model and fine-tuned all parameters from end to end. In practice, we only use the output from the [CLS] token as the representation of the whole sequence. Thus, the entire fine-tuning BERT architecture for the classification task is shown in Figure 1. A simple SoftMax classifier is added to the top of the model to predict the probability of label c shown in Equation 1. Where W is the task-specific parameter matrix. We fine-tune all the parameters from BERT as well as W jointly by maximizing the log-probability of the correct label.

$$p(c|h) = softmax(Wh) \qquad (1)$$

In this study, we use five pre-trained models as shown in Table 1. In the original paper, L represents the numbers of transformer layers (stacked encoder), H represents numbers of hidden embedding size, and A represents numbers of attention heads [8]. Smaller model architecture is using less parameters to train and can be used in limited computation resources. The number of parameters in every pretrained model shown in Table 2.

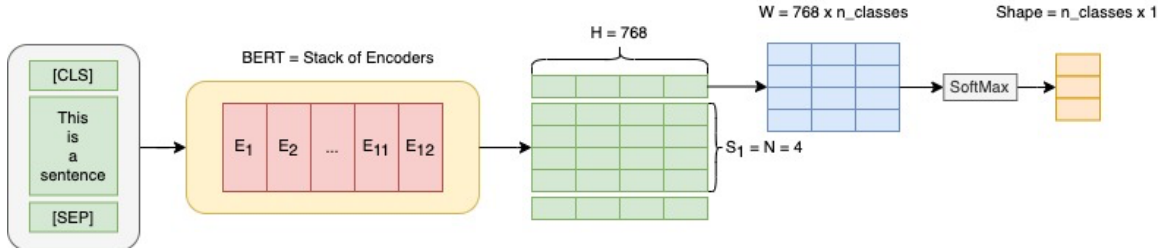

Figure 1: Fine-tuning BERT architecture for the classification task. We just use the [CLS] output token for classification along with some added Linear and SoftMax layers.

Table 1: Pre-trained BERT models are used. We only focus on six models: Tiny (L=2, H=128), Mini (L=4, H=256), Small (L=4, H=512), Medium (L=8, H=512), and Base (L=12, H=768).

|      | H=128     | H=256     | H=512       | H=768     |
|------|-----------|-----------|-------------|-----------|
| L=2  | BERT-Tiny | -         | -           | -         |
| L=4  | -         | BERT-Mini | BERT-Small  | -         |
| L=8  | -         | -         | BERT-Medium | -         |
| L=12 | -         | -         | -           | BERT-Base |

Table 2: The number of parameters on the pre-trained BERT model.

| Model | Parameters (Millions) |
|---|---|
| BERT-Tiny | 4.4 |
| BERT-Mini | 11.3 |
| BERT-Small | 29.1 |
| BERT-Medium | 41.7 |
| BERT-Base | 110.1 |
| BERT-Large | 340 |

The optimal hyperparameter values are task-specific. We use Adam with $\beta_1 = 0.9$ and $\beta_2 = 0.999$. The base learning rate is 1e-4 and the warm-up proportion is 0.1. We empirically set the max number of the epoch to 4 and save the best model on the validation set for testing.

### 3.3 Spark NLP

Due to the popularity of NLP in recent years, many NLP library have been developed, such as Natural Language Toolkit (NLTK) [39], SpaCy[3], TextBlob[4],Gensim[5], FastText [40], [41], and Stanford Core NLP [42]. Some of these are only optimized to work on a single node machine and not designed for distributed environments or parallel computing. In addition, recent deep learning models such as BERT have made significant changes to NLP because they can be fine-tuned and reused without major computational effort. A new library, Spark NLP, was introduced to meet the need for scalable, high-performance, and high-accuracy text processing. Spark NLP is an open-source library built on top of Apache Spark and Spark ML [17]. Apache Spark is a component of the Hadoop ecosystem, a favorite big data platform because of its ability to process streaming data.

In this study, we used text processing and word embedding from the BERT pre-trained model to build a text classification model in Spark NLP. Each stage in the Spark NLP is implemented in a pipeline as a sequence, as shown in Figure 2. Each resulting output is directed to the next stage as input. This means that the DataFrame (DF) input will be changed as it passes through each stage. First, DF is fed to DocumentAssembler() to generate document fields as starting points in Spark NLP. Then the document column is inserted into SenteceDectector() to be split into an array of sentences and generate a sentence column. The sentence column is inserted into Tokenizer() to generate a word token for the entire sentence and generate the token column. A token column is fed to BertEmbeddings() to convert the token into a vector representation. We use the BERT pre-trained model as explained in the previous section. Finally, we use Sentence Embeddings() to train a model.

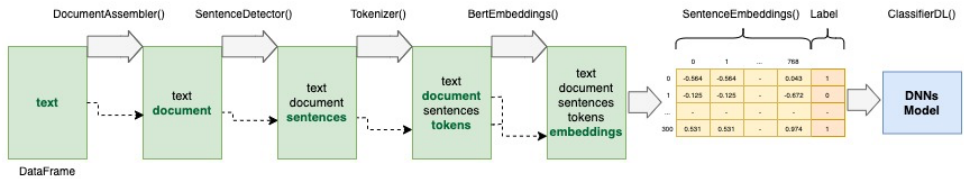

Figure 2: Spark NLP pipeline as a sequence for text classification. Each annotator applied adds a new column to a DataFrame that is fed into the pipeline.

---

[3] https://spacy.io.
[4] https://textblob.readthedocs.io/en/dev.
[5] https://radimrehurek.com/gensim.

To create a classifier in Spark NLP, we use ClassifierDL. ClassifierDL is a multi-class text classifier in Spark NLP, and it uses various text embeddings as an input for text classifications. The ClassifierDL uses a deep learning model (DNNs) built inside TensorFlow[6].The classification process is carried out after going through the text processing stages above. We will train each pre-trained model for 4 epochs with a batch size of 32 and a learning rate of 1e-4. Spark NLP will write the training logs to annotator_logs folder in our directory.

## 4  RESULT AND DISCUSSION

Our experiment compares basic BERT without Spark NLP and uses pipelines in Spark NLP to classify large text data. We run all models in Google Colab[7] containing one 1 GPU Tesla P100 16 GB and 27.4GB RAM. For the experiment environment, we use Python version 3.8, Spark NLP version 2.7.5, Apache Spark version 2.3.0, OpenJDK version 1.8.0_292, and TensorFlow version 2.4.1. We measure the resource needed when running all five pre-trained models in Table 1. In addition, we also measure the accuracy performance of the model on the testing data. We observe the GPU and RAM resources used during the training process. To track and observe during the model training process, we use the Weights & Biases[8] library running in our environment.

The first experiment is performed using the BERT model without Spark NLP. The first computing resources test results are shown in Figure 3. The BERT-Large model cannot run in our environment because it needs a higher specification than the one GPU we use in our experiment. According to the comparison shown in Figure 3, BERT-Base takes the most resources and needs a longer time to complete the training, and BERT-Tiny requires the least number of resources and completes the training the fastest. The result shows the bigger the model, the greater the GPU usage and memory allocation needed. Moreover, the bigger the model, the longer time it takes to complete model training.

In the next experiment, we used the pipeline from Spark NLP and added the embedding from the pre-trained BERT model to generate the word embedding. Model development using Spark NLP pipeline is relatively easy and fast compared to the BERT model from scratch. The results of testing computation resources when using Spark NLP are shown in Figure 4. The result is similar to the previous experiment, the bigger the model, the greater the GPU usage and memory allocation are needed. And the bigger the model is, the longer time to complete model training. This is because large models have a larger number of parameters to fine-tune. From the computation resource comparison shown in Figure 3 and Figure 4, we can see that BERT without Spark NLP needs the least number of resources and can complete the training much faster than BERT without Spark NLP.

BERT without Spark NLP and BERT with Spark NLP gives different results, as shown in Table 3. The highest accuracy of BERT without Spark NLP is 0.9253 by BERT-Base. This is not much different from the BERT-Small of 0.9213. Meanwhile, the lowest accuracy when using BERT-tiny is 0.9104. Thus, we did not see a significant increase in accuracy across the pre-trained BERT models without the Spark NLP with an average accuracy of 0.9187. The fastest training time using BERT-Tiny is 11 minutes. Meanwhile, the longest training time when using BERT-Base shows more than 1 hour. The average training time using BERT without spark NLP is 25 minutes. Similar to the previous experiment, the lowest accuracy was when using BERT-tiny at 0.8444, while the highest accuracy was obtained using BERT-base at 0.8665. The different results showed that accuracy continues to increase when using all pre-trained models using Spark NLP pipeline. Except for the BERT-Base medium, smaller than the BERT-Small. The average accuracy obtained for BERT with spark NLP is 0.8665. The average training time using BERT with spark NLP is 9 minutes.

---

[6] https://www.tensorflow.org.
[7] https://colab.research.google.com.
[8] https://wandb.ai.

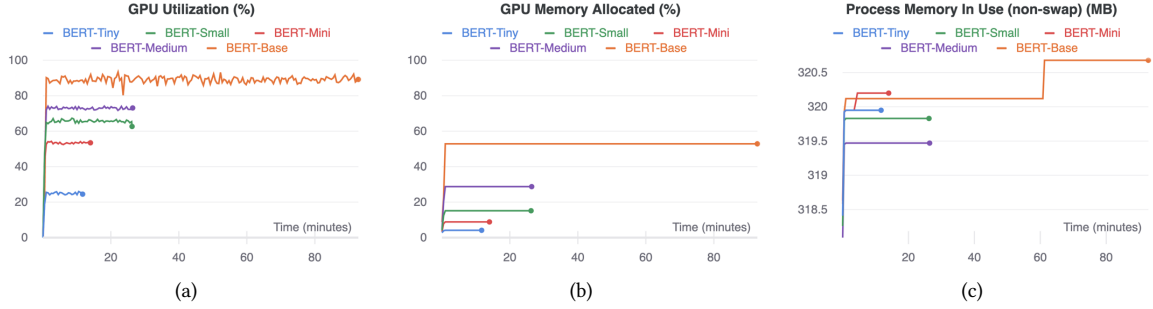

Figure 3: The computational resources used during the training use the BERT without Spark NLP pipeline. (a) GPU utilization, (b) GPU memory allocated, and (c) process memory in use.

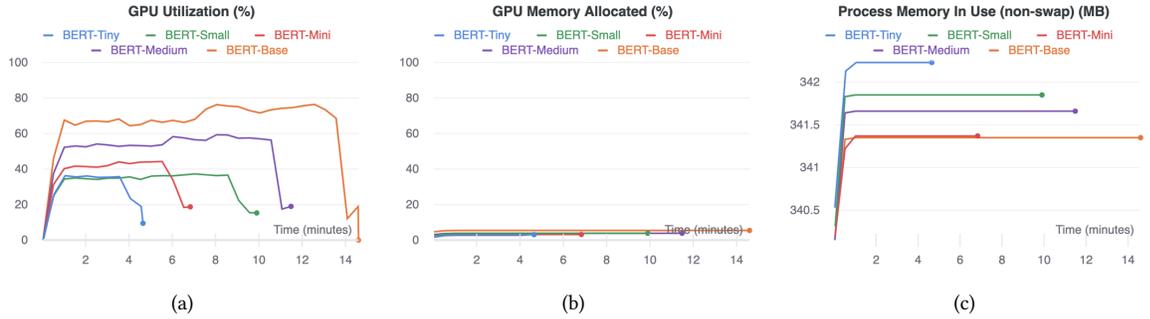

Figure 4: The computational resources used during the training use BERT with Spark NLP pipeline. (a) GPU utilization, (b) GPU memory allocated, and (c) process memory in use.

Table 3: Comparison of accuracy and computation time during the training process between the BERT without Spark NLP and BERT with Spark NLP pipelines. We also calculate the reduction in accuracy and computation time (in percent) to determine the effectiveness of the proposed pipeline.

|  | BERT | | BERT + Spark NLP | | Decrease in | Decrease in |
|---|---|---|---|---|---|---|
|  | Accuracy | Wall Time | Accuracy | Wall Time | Accuracy (%) | Time (%) |
| BERT-Tiny | 0.9104 | 00:11:41 | 0.8444 | 00:04:36 | 7.2 | 60.6 |
| BERT-Mini | 0.9168 | 00:13:59 | 0.8567 | 00:06:49 | 6.6 | 51.3 |
| BERT-Small | 0.9213 | 00:26:13 | 0.8714 | 00:10:17 | 5.4 | 60.8 |
| BERT-Medium | 0.9199 | 00:26:24 | 0.8710 | 00:11:28 | 5.3 | 56.6 |
| BERT-Base | 0.9253 | 01:38:12 | 0.8893 | 00:14:35 | 3.9 | 85.1 |
| Average | 0.9187 | 00:35:18 | 0.8665 | 00:09:33 | 5.7 | 62.9 |

The results of all experiments show that, BERT without Spark NLP gives higher accuracy rate compared to BERT with Spark NLP on all pre-trained models. But BERT with Spark NLP has advantages in efficiency. As shown in Table 3, BERT with Spark NLP gives good accuracy but it takes less time to complete the task. A significant decrease in computation time when using BERT with Spark NLP by 62.9% with a decrease in accuracy of 5.7%. Even though using Spark NLP the RAM resources used is much higher, we can see the efficiency of this method.

## 5 CONCLUSION

The BERT model is a good model to do large scale NLP tasks such as news classification. The larger the model gives higher accuracy, but it will take more time to complete the task. The bigger the dataset we use to train and test the model, it will affect the time it takes to complete the task. Using Spark NLP gives us advantages when we want to use a BERT-Large model and process large amounts of data. In this study we found that using BERT with Spark NLP is more efficient then using BERT without Spark NLP. Using BERT with Spark NLP, the drop accuracy average is 5.7% and the training time drop average is 62.9% compared to BERT without Spark NLP. In the near future, we plan to expand and improve our framework by exploring more architectures and pre-trained models to improve classification performance and computational resources. Furthermore, we wanted to explore the effects of text preprocessing prior to training.